\documentclass[10pt,twocolumn,letterpaper]{article}

\usepackage[pagenumbers]{cvpr} 

\usepackage{times}
\usepackage{epsfig}
\usepackage{graphicx}
\usepackage{amsmath}
\usepackage{amssymb}

\usepackage[caption = false]{subfig}
\usepackage{comment}
\usepackage{color}

\usepackage{dsfont}
\usepackage{multirow}
\usepackage{cite}

\usepackage{verbatim}
\usepackage{slashbox}
\usepackage{array}
\usepackage{booktabs, caption, makecell}

\usepackage{algorithm}
\usepackage{algpseudocode}%
\usepackage[accsupp]{axessibility} 

\usepackage[pagebackref=true,breaklinks=true,letterpaper=true,colorlinks,bookmarks=false]{hyperref}

\newcolumntype{L}[1]{>{\raggedright\arraybackslash}m{#1}}
\newcolumntype{C}[1]{>{\centering\arraybackslash}m{#1}}
\newcolumntype{R}[1]{>{\raggedleft\arraybackslash}m{#1}}
\newcolumntype{+}{>{\global\let\currentrowstyle\relax}}
\newcolumntype{^}{>{\currentrowstyle}}

\newcommand{\bfr}{\ensuremath{{\mathbf{r}}}}

\newcommand{\bfc}{\ensuremath{{\mathbf{c}}}}
\newcommand{\bfu}{\ensuremath{{\mathbf{u}}}}
\newcommand{\bfv}{\ensuremath{{\mathbf{v}}}}
\newcommand{\bfU}{\ensuremath{{\mathbf{U}}}}

\newcommand{\bfA}{\ensuremath{{\mathbf{A}}}}
\newcommand{\bfV}{\ensuremath{{\mathbf{V}}}}

\usepackage[capitalize]{cleveref}
\crefname{section}{Sec.}{Secs.}
\Crefname{section}{Section}{Sections}
\Crefname{table}{Table}{Tables}
\crefname{table}{Tab.}{Tabs.}

\def\cvprPaperID{6936} 
\def\confName{CVPR}
\def\confYear{2022}

\begin{document}

\title{Eigencontours: Novel Contour Descriptors Based on Low-Rank Approximation}

\author{Wonhui Park\\
Korea University\\
{\tt\small whpark@mcl.korea.ac.kr}
\and
Dongkwon Jin\\
Korea University\\
{\tt\small dongkwonjin@mcl.korea.ac.kr}
\and
Chang-Su Kim\\
Korea University\\
{\tt\small changsukim@korea.ac.kr}
}
\maketitle

\begin{abstract}
    Novel contour descriptors, called eigencontours, based on low-rank approximation are proposed in this paper. First, we construct a contour matrix containing all object boundaries in a training set. Second, we decompose the contour matrix into eigencontours via the best rank-$M$ approximation. Third, we represent an object boundary by a linear combination of the $M$ eigencontours. We also incorporate the eigencontours into an instance segmentation framework. Experimental results demonstrate that the proposed eigencontours can represent object boundaries more effectively and more efficiently than existing  descriptors in a low-dimensional space. Furthermore, the proposed algorithm yields meaningful performances on instance segmentation datasets.

\end{abstract}

\section{Introduction}

Contour is one of the most important object descriptors, along with texture and color. The boundary of an object in an image is encoded in contour description, which is useful in various applications, such as image retrieval \cite{chuang1996,zhang2002,zhang2004review}, recognition \cite{mokhtarian1992,shotton2008, xu2012}, and segmentation \cite{xie2020,xu2019, maninis2017, zhen2020joint, peng2020}. It is desirable to represent object boundaries compactly, as well as faithfully, but it is challenging to design such contour descriptors due to the diversity and complexity of object shapes.

Early contour descriptors were developed mainly for image retrieval \cite{mokhtarian1992,chuang1996,zhang2002,zhang2004review}. An object contour can be simply represented based on the area, circularity, and/or eccentricity of the object \cite{young1974}. For more precise description, there are several approaches, including shape signature \cite{davies2004,van1991,xie2020}, structural analysis \cite{freeman1978,perez1994,dierckx1995,cinque1998,xu2019}, spectral analysis \cite{chuang1996,zhang2002}, and curvature scale space (CSS) \cite{mokhtarian1992,dudek1997}.

Recently, contour descriptors have been incorporated into deep-learning-based object detection, tracking, and segmentation systems. In \cite{zhou2019bottom}, bounding boxes are replaced by polygons to enclose objects more tightly. In \cite{xin2019fast}, ellipse fitting is done to produce a rotated box of a target object to be tracked. For instance segmentation, contour-based techniques have been proposed that represent pixelwise masks by contour descriptors based on shape signature \cite{xie2020} or polynomial fitting \cite{xu2019}. Even though these descriptors can localize an object effectively, they may fail to reconstruct the object boundary faithfully. Also, they consider the structural information of an individual object only, without exploiting the shape correlation between different objects.

\begin{figure}[t]
  \centering
  \includegraphics[width=1\linewidth]{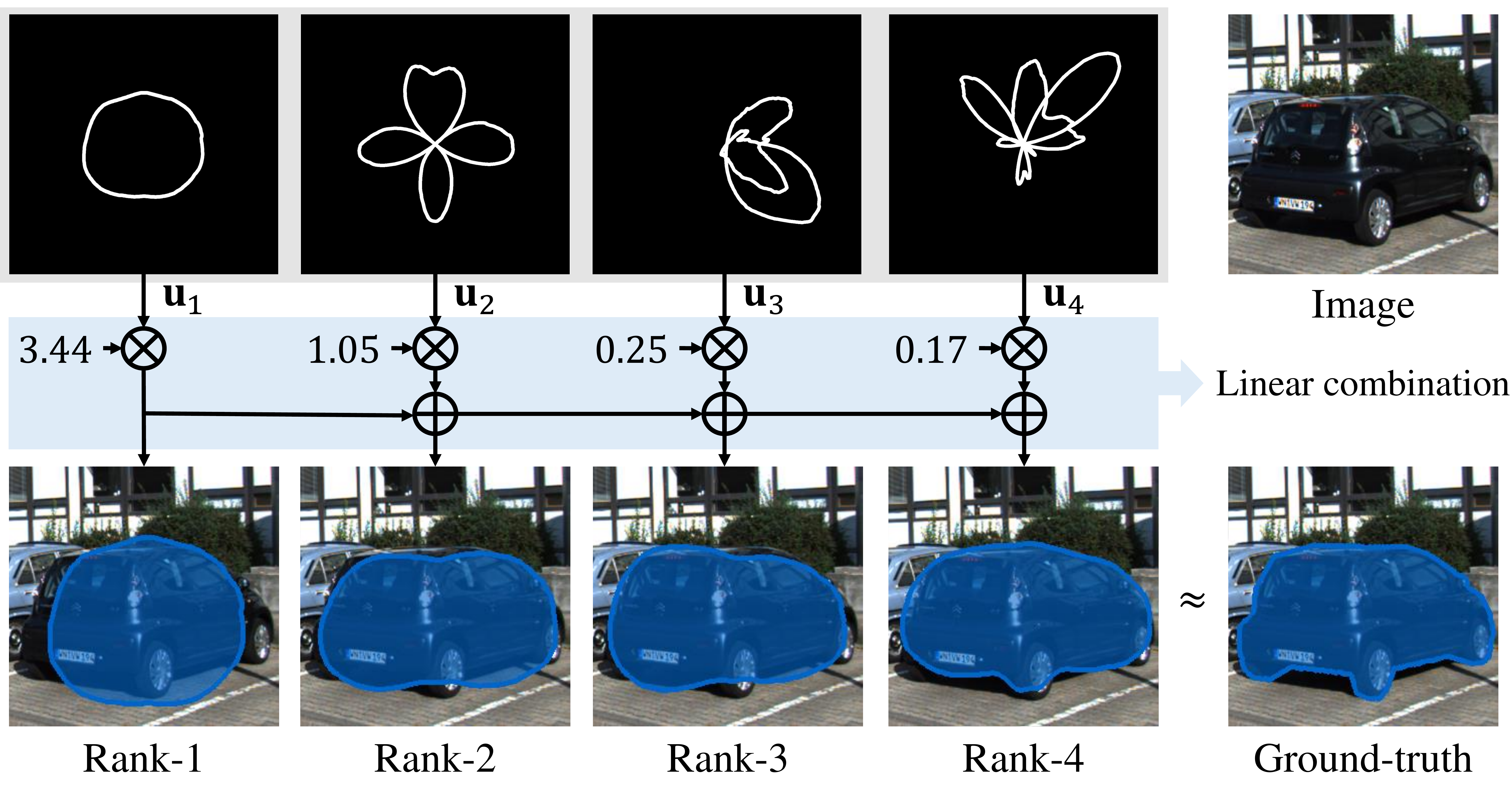}
  \caption{Illustration of the eigencontour representation. The boundary of a vehicle is represented by a linear combination of four eigencontours: $\bfu_1$, $\bfu_2$, $\bfu_3$ and $\bfu_4$. First, $\bfu_1$ approximates the object boundary roughly. Next, $\bfu_2$ is used to refine the boundary by adjusting top and bottom parts, as well as front and rear ones. To reconstruct more complex parts, such as wheels and bumper, $\bfu_3$ and $\bfu_4$ should be used as well. These eigencontours were determined by analyzing the boundaries of all objects in the `car' category in the KINS dataset \cite{qi2019kins}.}
  \label{fig:Intro_fig}
  \vspace*{-0.3cm}
\end{figure}

In this paper, we propose novel contour descriptors, called \textit{eigencontours}, based on low-rank approximation. First, we construct a contour matrix containing all object boundaries in a training set. Second, we decompose the contour matrix into eigencontours, based on the best rank-$M$ approximation of singular value decomposition (SVD) \cite{y2015SVD}. Then, each contour is represented by a linear combination of the $M$ eigencontours, as illustrated in Figure \ref{fig:Intro_fig}. Also, we incorporate the eigencontours into an instance segmentation framework. Experimental results demonstrate that the proposed eigencontours can represent object boundaries more effectively and more efficiently than the existing contour descriptors \cite{xie2020,xu2019}. Moreover, utilizing the existing framework of YOLOv3 \cite{redmon2018}, the proposed algorithm yields promising instance segmentation performances on various datasets --- KINS \cite{qi2019kins}, SBD \cite{hariharan2011}, and COCO2017 \cite{lin2014}.

This work has the following contributions:
\begin{itemize}
\itemsep0mm
\item We propose the notion of eigencontours --- data-driven contour descriptors based on SVD --- to represent object boundaries as faithfully as possible with a limited number of coefficients.
\item The proposed algorithm can represent object boundaries more effectively and more efficiently than the existing contour descriptors.
\item The proposed algorithm outperforms conventional contour-based techniques in instance segmentation.
\end{itemize}

\section{Related Work}\label{sec:related}
The goal of contour description is to represent the boundary of an object in an image compactly and faithfully. Simple contour descriptors are based on the area, circularity, and/or eccentricity of an object \cite{young1974}, and basic geometric shapes, such as rectangles and ellipses, can be also used. However, these simple descriptors cannot preserve the original shape of an object faithfully \cite{zhang2020mask}, \cite{shen2021dct}.
For more sophisticated description, there are four types of approaches: shape signature \cite{davies2004,van1991,xie2020}, structural analysis \cite{freeman1978,perez1994,cinque1998,xu2019}, spectral analysis \cite{chuang1996,zhang2002}, and CSS \cite{mokhtarian1992,dudek1997}.
First, a shape signature is a one-dimensional function derived from the boundary coordinates of an object. For example, a polar coordinate system is set up with respect to the centroid of an object. Then, the object boundary is represented by the $(r, \theta)$ graph, called the centroidal profile \cite{davies2004}. Also, an object shape can be represented by the angle between the tangent vector at each contour point and the $x$-axis \cite{van1991}. Second, structural methods divide an object boundary into segments and approximate each segment to encode the whole boundary. In \cite{freeman1978}, the boundary is represented by a sequence of unit vectors with a few possible directions. In \cite{perez1994}, polygonal approximation is performed to globally minimize the errors from an approximated polygon to the original boundary. In \cite{cinque1998}, segments of an object contour are represented by cubic polynomials. Third, in spectral methods, boundary coordinates are transformed to a spectral domain. In \cite{chuang1996}, a wavelet transform is used for contour description.
In \cite{zhang2002}, the Fourier descriptors are derived from the Fourier series of centroidal profiles. Fourth, in CSS \cite{mokhtarian1992}, a boundary is smoothed by a Gaussian filter with a varying standard deviation. Then, the boundary is represented by the curvature zero-crossing points of the smoothed curve at each standard deviation.

\begin{figure}[t]
  \centering
  \includegraphics[width=1\linewidth]{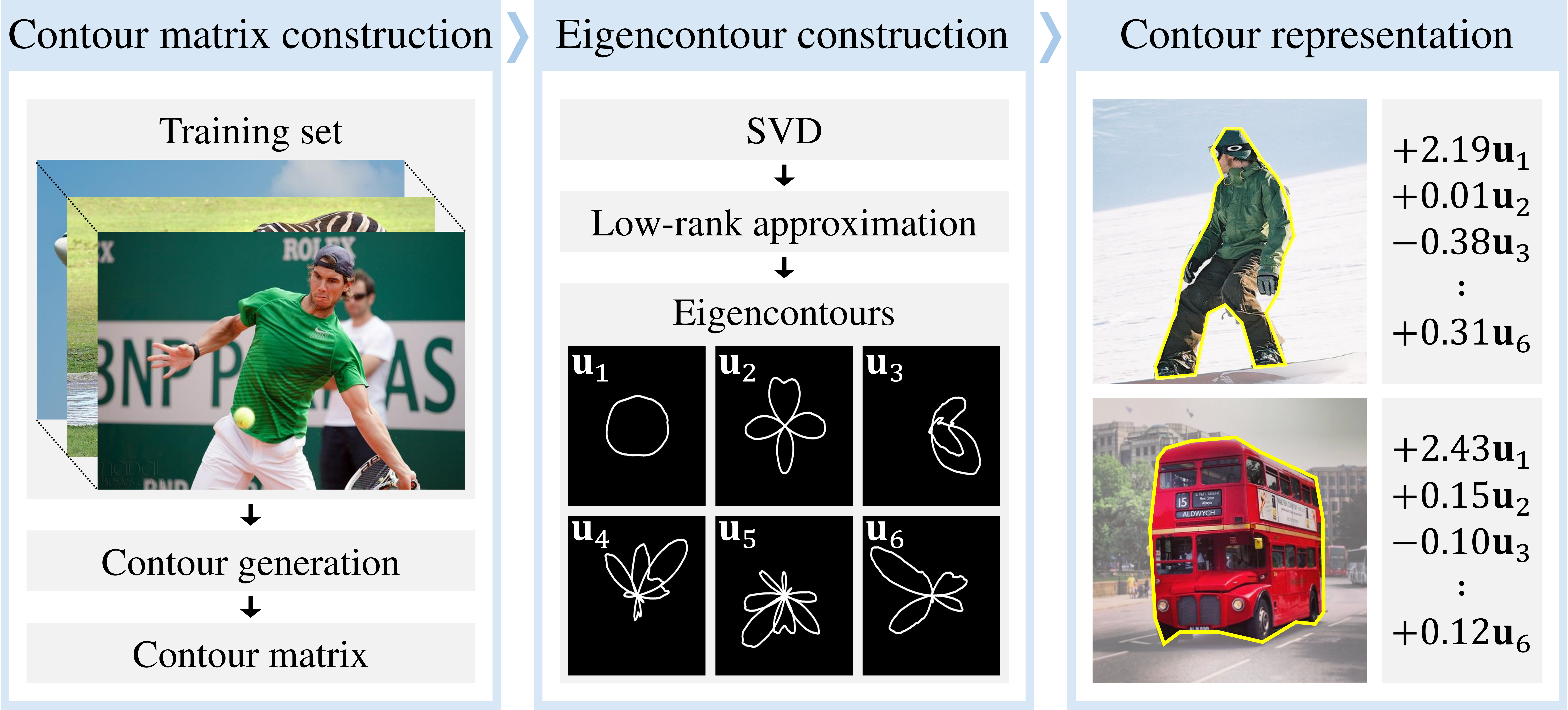}
  \setlength{\abovecaptionskip}{-0.3cm}
  \caption{Overview of the proposed algorithm.}
  \vspace*{-0.1cm}
  \label{fig:overview_fig}
\end{figure}

Recently, attempts have been made to improve the performances of deep-learning-based vision systems. In \cite{zhou2019bottom}, a bounding box for object detection is replaced by an octagon to enclose an object more tightly via polygonal approximation. In \cite{xin2019fast}, a rotated box for a target object is determined based on ellipse fitting, in order to cope with object deformation in a visual tracking system. For instance segmentation, contour-based approaches \cite{xu2019,xie2020} have been developed, which reformulate the pixelwise classification task as the boundary regression of an object. To this end, these methods encode segmentation masks into contour descriptors. In \cite{xie2020}, centroidal profiles are used to describe object boundaries.
In \cite{xu2019}, each segment of a boundary is represented by a few coefficients based on polynomial fitting. Although these methods are computationally efficient for localizing object instances, they often fail to reconstruct the boundaries of the object shapes faithfully.

The proposed algorithm aims to represent an object boundary as faithfully as possible by employing as few coefficients as possible. To this end, we develop eigencontours based on the best low-rank approximation property of SVD.

\section{Proposed Algorithm}
Instead of deriving contour descriptors based on prior assumptions on object boundaries, such as rectangular, elliptical, or polynomial models, we develop eigencontours by analyzing boundary data in a training set. In this sense, the proposed eigencontours are data-driven descriptors. Figure~\ref{fig:overview_fig} is an overview of the proposed algorithm.
First, we compose a contour matrix, containing all object boundaries in a training set. Second, we approximate the matrix, by performing the best rank-$M$ approximation, to determine $M$ eigencontours. Third, we represent an object boundary by a linear combination of the $M$ eigencontours.

\subsection{Mathematical Formulation}
\label{ssec:formulation}

SVD and principal component analysis (PCA) are used in various fields to achieve dimensionality reduction and represent data concisely \cite{Linear2012,y2015SVD,jin2022}. In this paper, we use SVD to represent object boundaries compactly and reliably. More specifically, we adopt a data-driven approach to exploit the distribution of object contours in a training set, instead of performing curve fitting \cite{cinque1998} or Fourier analysis \cite{zhang2002}, in order to represent object boundaries efficiently in a low-dimensional space.

\begin{figure}[t]
  \centering
  \includegraphics[width=1\linewidth]{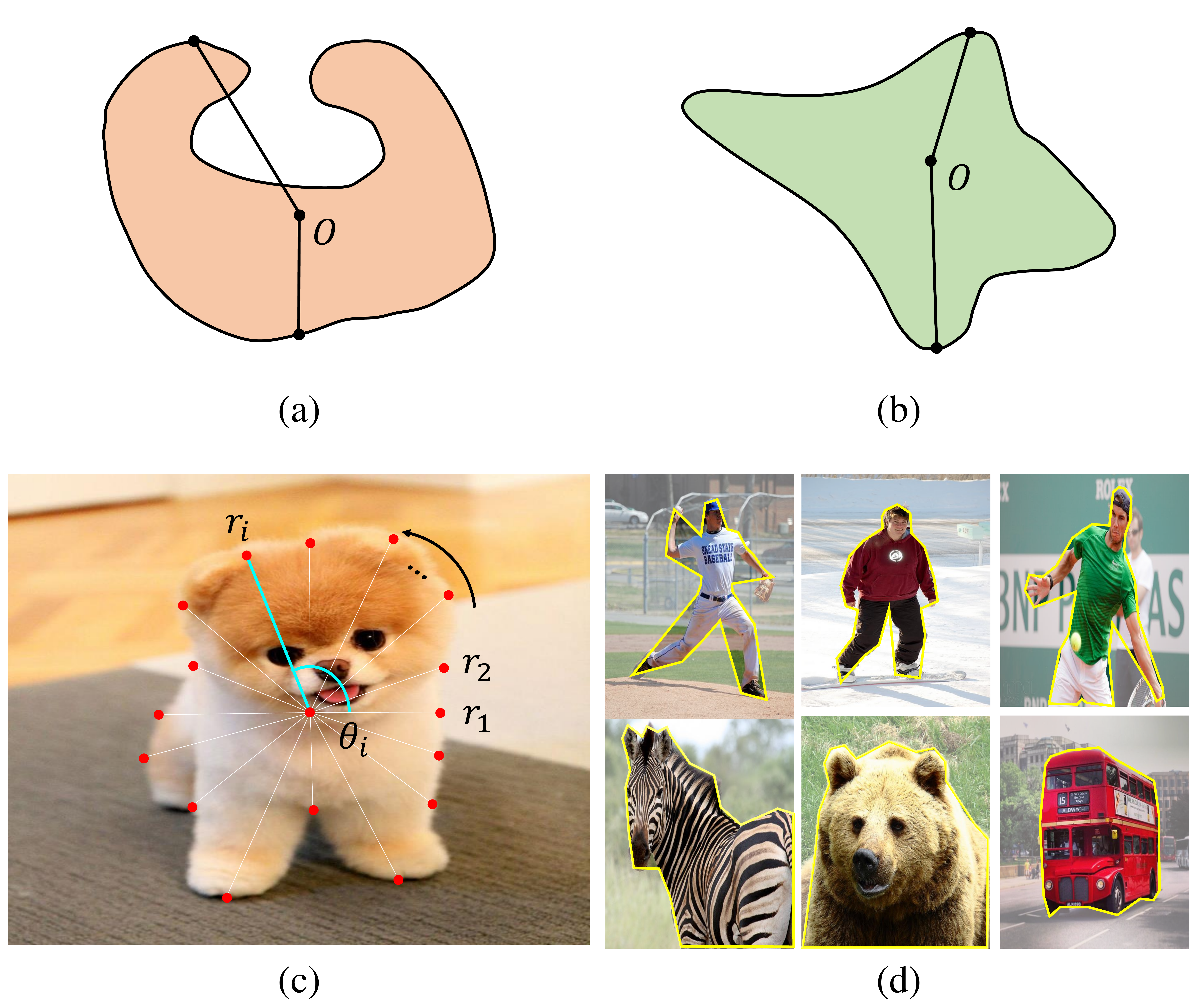}
  \vspace*{-0.5cm}
  \caption{In (a), the set (or shape) is not star-convex because there are line segments not wholly contained in the set. In (b), it is star-convex because the segment from the center $O$ to any point in the set is contained in the set. In (c), a star-convex contour is represented by polar coordinates. More examples of star-convex contours are in (d).}
  \label{fig:starconvex}
\end{figure}

\vspace*{-0.4cm}
\noindent\textbf{Star-convex contour generation:}
There is a tradeoff between accuracy and simplicity of a contour representation scheme: an accurate representation yields a high-dimen\-sional feature vector, while too simple a representation cannot describe complicated boundaries precisely. To strike a good balance, we adopt the star-convexity assumption of object shapes. A regional set (or shape) is star-convex \cite{stanek1977characterization} if it contains a point such that the line segment from the point to any point in the set is contained in the set. Then, a star-convex contour is defined as the set of boundary points of a star-convex set. For example, Figure \ref{fig:starconvex}(a) is not a star-convex contour, but Figure \ref{fig:starconvex}(b) is a star-convex one.

To represent star-convex contours, we use centroidal profiles \cite{davies2004}. Given an object shape, we find the inner-center, which is the center of the circle of the maximum size wholly contained in the shape, as done in \cite{xu2019}. Then, with respect to the inner center, we describe the boundary using polar coordinates $(r_i, \theta_i)$, $i=1, 2, \ldots, N$. The angular coordinates $\theta_i$ are sampled uniformly, so only the radial coordinates are recorded to represent the contour
\begin{equation}
\bfr=[r_1, r_2, \ldots, r_N]^\top.
\label{eq:contour_def}
\end{equation}
As in Figure~\ref{fig:starconvex}(c), $r_i$ is set to be the distance of the farthest object point from the center along the $\theta_i$-axis. By construction, $\bfr$ describes a star-convex contour.

Figure \ref{fig:starconvex}(d) shows more star-convex contours. With the infinite sampling $N=\infty$, a star-convex contour is guaranteed to enclose all object points, since it is the boundary of the star-convex hull of the object. However, with a finite $N$, the star-convex contour may miss some object points, as well as include some non-object points. However, we see that the contours in Figure \ref{fig:starconvex}(d) represent object shapes quite faithfully.

\vspace*{0.1cm}
\noindent\textbf{Eigencontour space:}
In general, object shapes are well structured and thus highly correlated to one another, especially between objects in the same class. By exploiting this structural relationship using big data, we design effective contour descriptors. Specifically, we first construct a star-convex contour matrix ${\mathbf A}=[\bfr_1, \bfr_2, \cdots, \bfr_L]$ from $L$ training objects. Then, we perform SVD of the matrix $\mathbf A$,
\begin{equation}\label{eq:svd}
    \textstyle
    \mathbf{A} = \mathbf{U} \mathbf{\Sigma} \mathbf{V}^\top
\end{equation}
where $\mathbf{U} = [\bfu_1, \cdots, \bfu_N]$ and $\bfV = [\bfv_1, \cdots, \bfv_L]$ are orthogonal matrices and $\mathbf{\Sigma}$ is a diagonal matrix, composed of singular values $\sigma_1 \geq \sigma_2 \geq \cdots \geq \sigma_r > 0$. It is known that
\begin{equation}\label{eq:rank-m}
    \bfA_M  = [\tilde{\bfr}_1, \cdots, \tilde{\bfr}_L] = \sigma_{1}{\mathbf u}_{1}{\mathbf v}^\top_{1} + \cdots + \sigma_{M}{\mathbf u}_{M}{\mathbf v}^\top_{M}
\end{equation}
is the best rank-$M$ approximation of $\bfA$ \cite{y2015SVD}.

In \eqref{eq:rank-m}, each approximate contour $\tilde{\bfr}_i$ is given by a linear combination of the first $M$ left singular vectors $\bfu_1, \cdots, \bfu_M$. In other words,
\begin{equation} \label{eq:x_approx}
\tilde{\bfr}_i = \bfU_M \bfc_i = [\bfu_1, \cdots, \bfu_M] \bfc_i.
\end{equation}
We refer to these vectors $\bfu_1, \cdots, \bfu_M$ as \textit{eigencontours}, and the space spanned by $\{\bfu_1, \cdots, \bfu_M\}$ as the \textit{eigencontour space}.

Given a contour $\bfr$, we project it onto the eigencontour space to obtain the low-rank approximation
\begin{equation}\label{eq:backward}
    \tilde{\bfr} = \bfU_M \bfc
\end{equation}
where the coefficient vector $\bfc$ is given by
\begin{equation}\label{eq:forward}
    \bfc = \bfU_M^\top \bfr.
\end{equation}
In \eqref{eq:forward}, an $N$-dimensional contour $\bfr$ is optimally approximated by an $M$-dimensional vector $\bfc$ in the eigencontour space, where $M < N$. Also, the approximate $\tilde{\bfr}$ can be reconstructed from $\bfc$ via \eqref{eq:backward}. Note that eigencontours may have negative elements. Thus, in rare cases, the approximate $\tilde{\bfr}$ has negative elements. In such cases, we truncate the negative elements to 0 to ensure the star-convexity of $\tilde{\bfr}$.

\vspace*{0.1cm}
\noindent\textbf{Clustering in eigencontour space:}
To discover typical contour patterns in a dataset, contour clustering can be performed. Instead of the original contour space of dimension $N$, contours can be grouped more effectively and more efficiently in the eigencontour space of dimension $M$. This is because the original space is transformed to the eigencontour space by an isometry $\bfU_M^\top$. Specifically, let $\tilde{\bfr}_1, \ldots, \tilde{\bfr}_L$ be object contours, which are approximated via \eqref{eq:x_approx}. Then, it can be easily shown that
\begin{equation}
\label{eq:clustering}
\| \tilde{\bfr}_i - \tilde{\bfr}_j\| =  \|\bfc_i - \bfc_j\|.
\end{equation}
In other words, the distances between contours in the original space are equal to those between the corresponding coefficient vectors in the eigencontour space. Hence, the clustering can be performed to yield the same results in both spaces, but it can be done more reliably and more efficiently in the eigencontour space because $M < N$. Note that, as the dimension of a space gets higher, clustering  becomes more difficult because of the curse of dimensionality~\cite{bellman1966dynamic}.

\begin{figure}[t]
  \centering
  \includegraphics[width=1\linewidth]{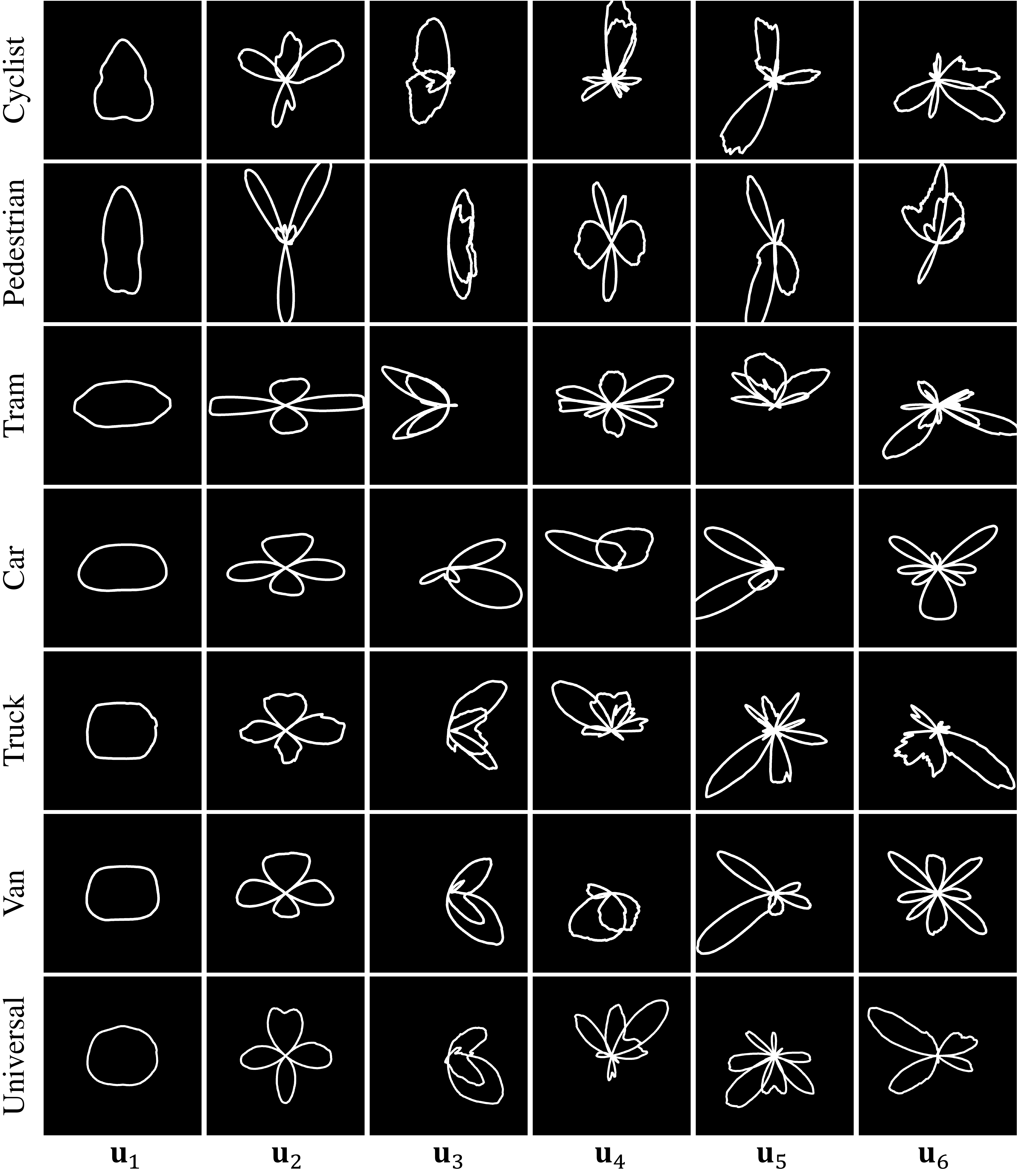}
  \vspace*{-0.5cm}
  \caption{The first six eigencontours $\bfu_1, \bfu_2, \ldots, \bfu_6$ for the KINS dataset. The top six rows show the eigencontours for separate object categories in KINS, while the bottom one shows those for the universal set of all instances in the six categories.}
  \label{fig:Eigenspace_fig}
  \vspace*{-0.2cm}
\end{figure}

\vspace*{0.1cm}
\noindent\textbf{Regression in eigencontour space:}
Furthermore, it is also beneficial to find object contours in the eigencontour space. A contour regressor can be designed to detect object boundaries in images. To detect a star-convex contour in \eqref{eq:contour_def} in the original space, we should regress $N$ variables. However, we can approximate all ground-truth contours of training objects using the first $M$ eigencontours and train a network to regress $M$ coefficients of $\bfc$ in \eqref{eq:forward} in the eigencontour space. This approach requires the regression of fewer variables. Hence, the regression network also needs fewer parameters and is more efficient in both training and inference stages. The efficacy of the regression in the eigencontour space is demonstrated in Sections \ref{ssec:assess} and \ref{ssec:ablation}.

\subsection{Examples and Analysis}

\noindent\textbf{Eigencontours:}
In this example, we use the KINS dataset \cite{qi2019kins}, the instances of which are divided into seven categories. We determine the eigencontours for the six categories of `cyclist,' `pedestrian,' `tram,' `car,' `truck,' and `van,' respectively, except for `misc' containing miscellaneous instances with unspecified classes. We also obtain the eigencontours for the universal set of all instances in the six categories. Each object boundary is represented by a 360-dimensional star-convex contour vector, by uniformly quantizing the 360-degree with an interval of $1^\circ$, \ie $N=360$.

\begin{figure}[t]
  \centering
  \includegraphics[width=1\linewidth]{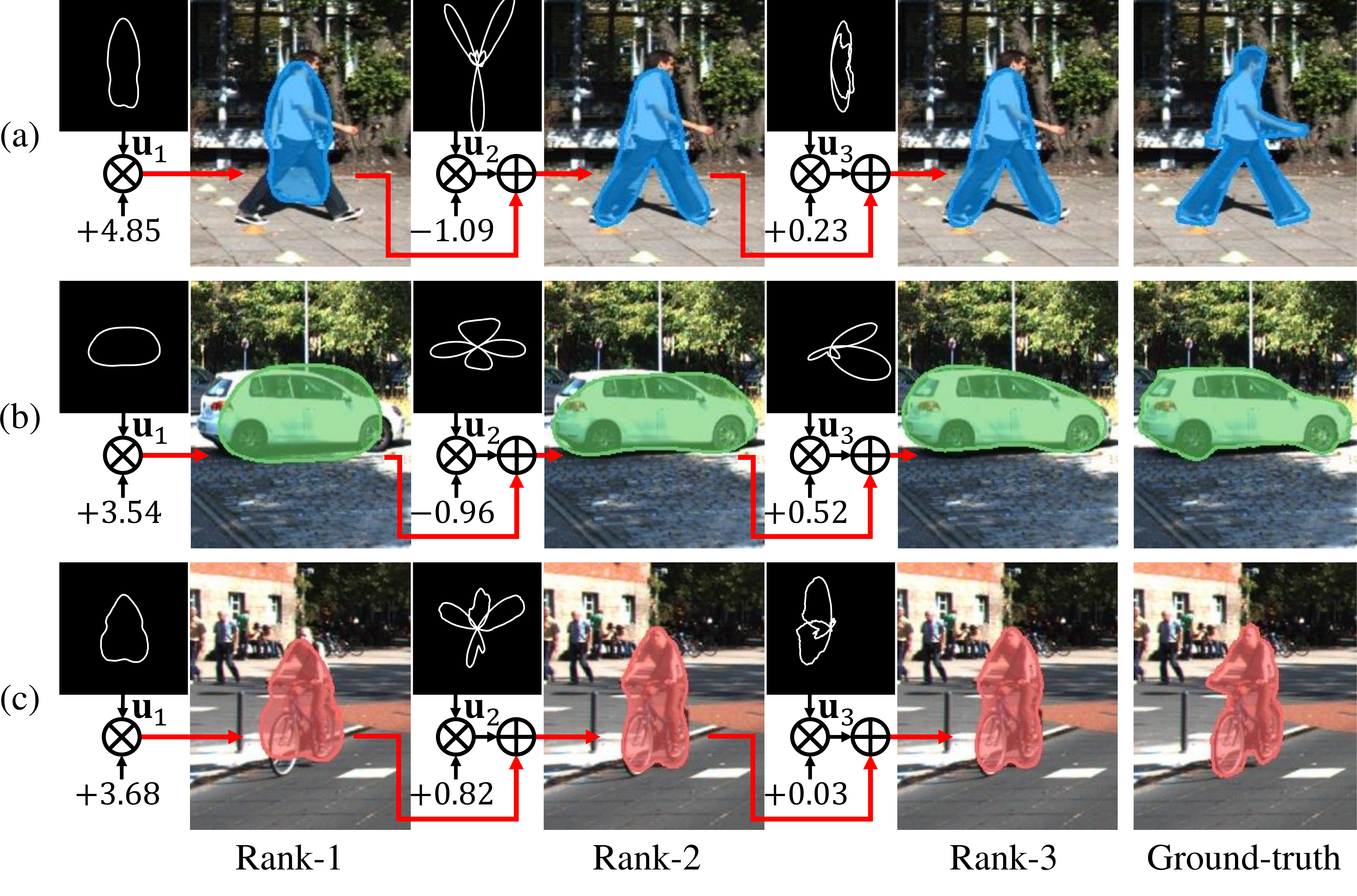}
  \vspace*{-0.5cm}
  \caption{Illustration of linear combination of eigencontours.}
  \label{fig:linear}
\end{figure}

Figure \ref{fig:Eigenspace_fig} shows the first six eigencontours $\bfu_1, \ldots, \bfu_6$. For each category, the first eigencontour $\bfu_1$ describes rough outlines of typical instances. For example, most pedestrians stand or walk on sidewalks, as implied by the vertical shape of $\bfu_1$ for `pedestrian.' By weighting $\bfu_1$, the size of the shape can be controlled. Next, $\bfu_2$ is more complicated to represent detailed parts of instances. For `pedestrian,' $\bfu_2$ is used to reconstruct a pair of legs, as shown in Figure \ref{fig:linear}(a). Also, $\bfu_2$ for `car'  generates a streamlined shape by refining the four sides of a car in Figure \ref{fig:linear}(b). The coefficient for $\bfu_2$ affects the horizontal and vertical sizes of the car.  Similarly, $\bfu_2$ for `cyclist' recovers bike wheels in Figure \ref{fig:linear}(c). In general, the coefficients for $\bfu_1$ and $\bfu_2$ are larger than those for the other eigencontours, and they are major factors for determining overall shapes. To represent those shapes more precisely, more eigencontours are required. Note that, for the three related categories of `car,' `truck,' and `van,' $\bfu_1$ and $\bfu_2$ are similar to one another. Also, $\bfu_1$ for the universal set is a round shape to describe various instances in different categories.

\begin{figure*}[t]

  \centering
  \includegraphics[width=0.9\linewidth]{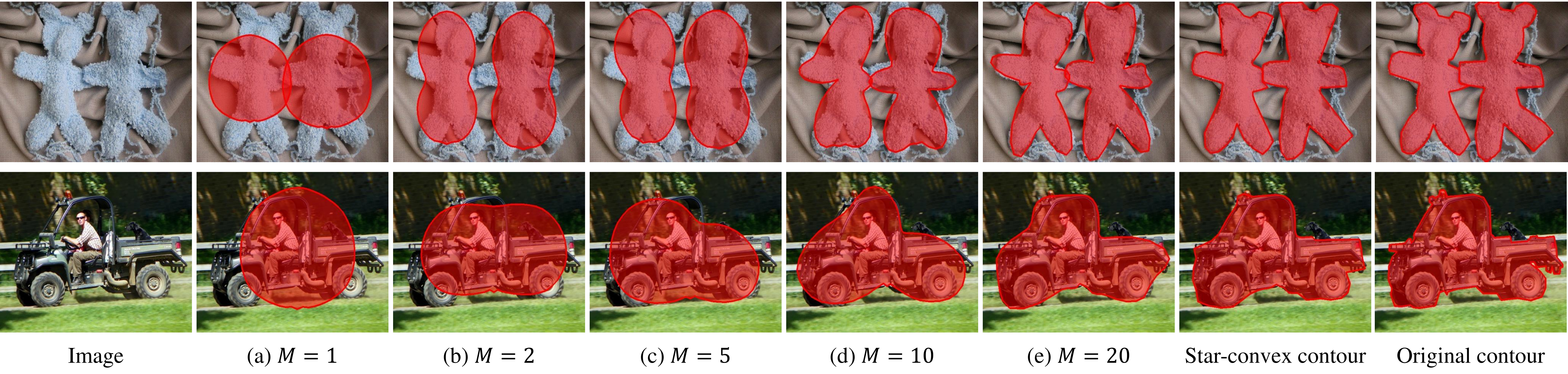}
  \caption{Object boundaries can be approximated using the first $M$ eigencontours. As $M$ gets larger, the rank-$M$ approximations get closer to the star-convex conversions of the original contours.}
  \vspace*{-0.2cm}
  \label{fig:recon_fig}
\end{figure*}

\begin{figure}[t]

  \centering
  \includegraphics[width=1\linewidth]{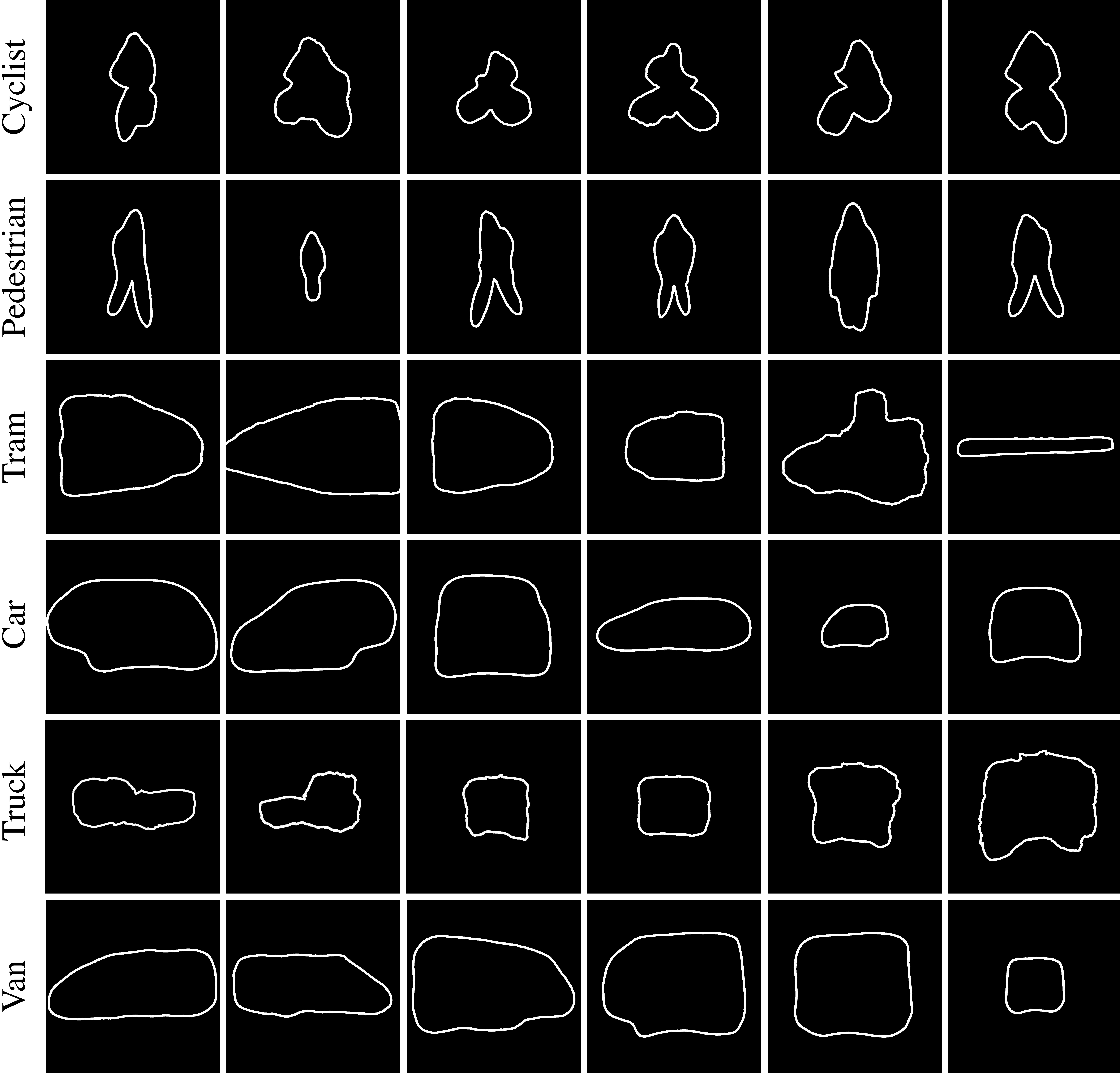}
    \setlength{\abovecaptionskip}{-0.1cm}
  \caption{Visualization of contour centroids in the $16$-dimensional eigencontour space, according to object categories. Although the centroids are determined by grouping training data in the lower-dimensional space, each centroid represents the structure of the corresponding object category faithfully.}
    \vspace{-0.2cm}
  \label{fig:clustering_fig}
\end{figure}

\begin{figure*}[t]
  \centering
  \includegraphics[width=1\linewidth]{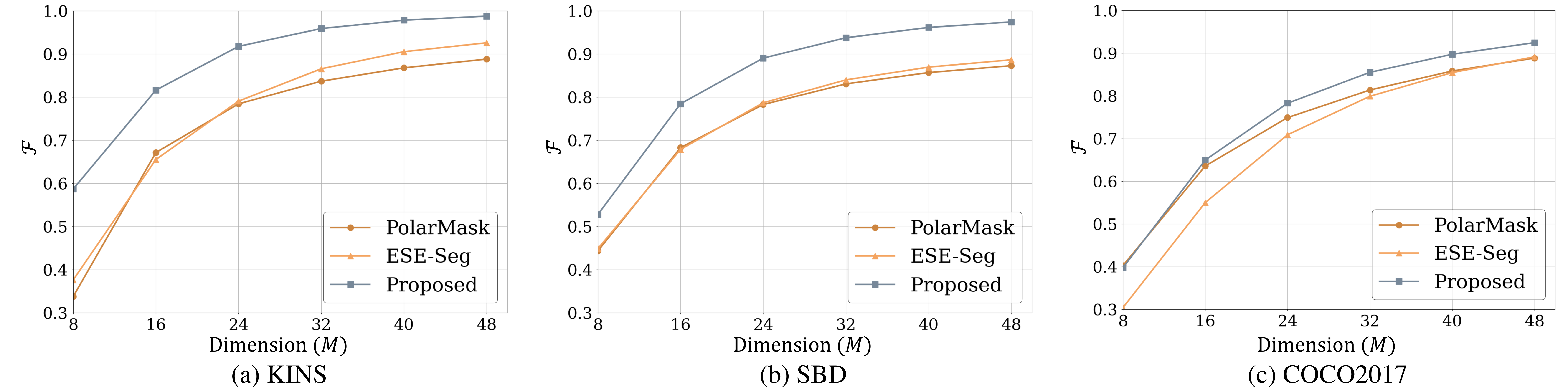}
  \setlength{\abovecaptionskip}{-0.2cm}
  \caption{The $\cal F$ score curves of the proposed eigencontours and the conventional contour descriptors in PolarMask  \cite{xie2020} and ESE-Seg  \cite{xu2019} according to the dimension $M$ of the descriptors.}
  \label{fig:graph_fig}
  \vspace*{0.1cm}
\end{figure*}

\vspace*{0.1cm}
\noindent\textbf{Rank-$M$ approximation:}
Figure \ref{fig:recon_fig} shows two object boundaries in the COCO2017 dataset \cite{lin2014} and their rank-$M$ approximations. In this test, the eigencontours are determined for all training instances in all categories. The rank-1 approximations are not good enough; they represent the overall sizes of the objects only. The rank-2 approximations better reconstruct object shapes, but only roughly. As $M$ gets larger, more faithful contours are restored. In this example, the objects have relatively complex shapes. Hence, to represent their boundaries well, the rank-20 approximations are required, which are almost identical to the 360-dimensional star-convex contours. Although they cannot reconstruct the original contours perfectly, it is not because of the low-rank approximation, but because of the star-convex conversion. Note that, compared to the 360-dimensional star-convex contours, the rank-20 approximations reduce the dimensionality by a factor of 18.

\vspace*{0.1cm}
\noindent\textbf{Clustering in eigencontour space:}
For each of the six categories in the KINS dataset, we cluster the object boundaries in the 16-dimensional eigencontour space ($M=16$) using the $K$-means algorithm, where $K$ is set to 100.
Figure~\ref{fig:clustering_fig} shows examples of contour centroids. We see that the centroids represent typical object shapes in the categories from different views. This indicates that eigencontours are effective not only for representing individual contours faithfully, but also for clustering contours into typical patterns in a lower-dimensional space.

\section{Experiments}

\subsection{Datasets}
\label{ssec:dataset}
We use three datasets: KINS, SBD, and COCO2017. All these datasets were approved by institutional review boards.

\vspace*{0.1cm}
\noindent\textbf{KINS \cite{qi2019kins}:}
It is a dataset for amodal instance segmentation, built on the KITTI dataset \cite{geiger2012}. It consists of 7,474 training and 7,517 test images. All instances are classified into seven categories, and an amodal segmentation mask is annotated for each instance.

\vspace*{0.1cm}
\noindent\textbf{SBD \cite{hariharan2011}:}
It is a semantic boundary dataset, re-annotated from the PASCAL VOC dataset \cite{everingham2010}. Its 11,355 images are split to 5,623 training and 5,732 validation images. All instances are classified into 20 object categories. Each instance is annotated with its shape boundary without holes.

\vspace*{0.1cm}
\noindent\textbf{COCO2017 \cite{lin2014}:}
It is a large dataset for various tasks, such as object detection and segmentation. It contains 118K training images, 5K validation images, and 41K test images. The instance segmentation masks for objects in 80 categories are provided.

\begin{figure*}[t]
\vspace{-0.5cm}

    \subfloat {\raisebox{1.6em}{\rotatebox[origin=t]{90}{\scriptsize Image}}}\hspace{-0.01cm}\,\!
    \subfloat {\includegraphics[width=1.68cm,height=1.4cm]{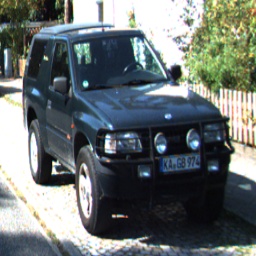}}\,\!\!
    \subfloat {\includegraphics[width=1.68cm,height=1.4cm]{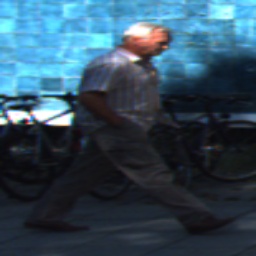}}\,\!\!
    \subfloat {\includegraphics[width=1.68cm,height=1.4cm]{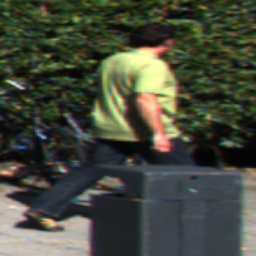}}\,\!\!
    \subfloat {\includegraphics[width=1.68cm,height=1.4cm]{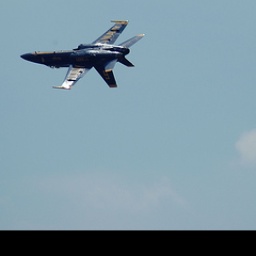}}\,\!\!
    \subfloat {\includegraphics[width=1.68cm,height=1.4cm]{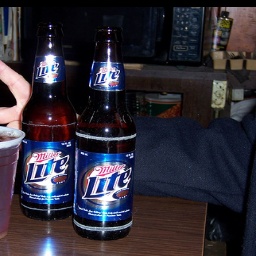}}\,\!\!
    \subfloat {\includegraphics[width=1.68cm,height=1.4cm]{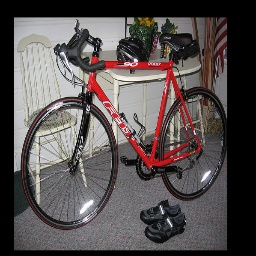}}\,\!\!
    \subfloat {\includegraphics[width=1.68cm,height=1.4cm]{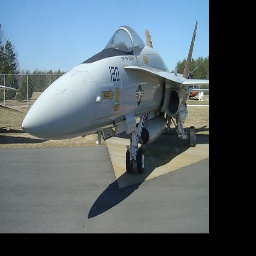}}\,\!\!
    \subfloat {\includegraphics[width=1.68cm,height=1.4cm]{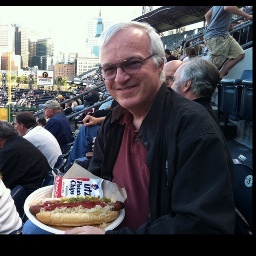}}\,\!\!
    \subfloat {\includegraphics[width=1.68cm,height=1.4cm]{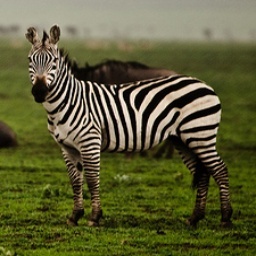}}\,\!\!
    \subfloat {\includegraphics[width=1.68cm,height=1.4cm]{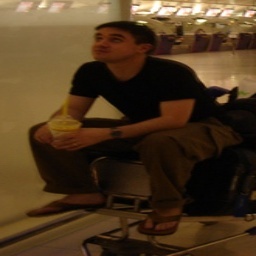}}\\[-5.2ex]

    \setcounter{subfigure}{-1}
    \subfloat {\raisebox{1.6em}{\rotatebox[origin=t]{90}{\scriptsize PolarMask \cite{xie2020}}}}\hspace{0.0cm}\,\!
    \subfloat {\includegraphics[width=1.68cm,height=1.4cm]{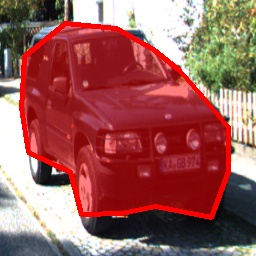}}\,\!\!
    \subfloat {\includegraphics[width=1.68cm,height=1.4cm]{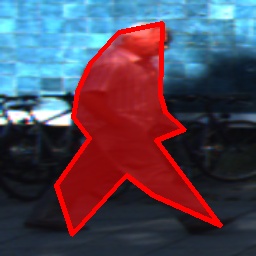}}\,\!\!
    \subfloat {\includegraphics[width=1.68cm,height=1.4cm]{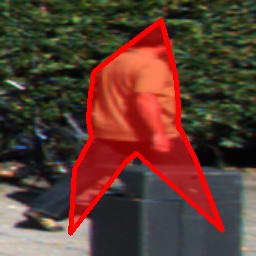}}\,\!\!
    \subfloat {\includegraphics[width=1.68cm,height=1.4cm]{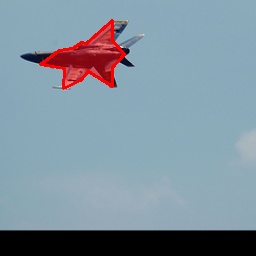}}\,\!\!
    \subfloat {\includegraphics[width=1.68cm,height=1.4cm]{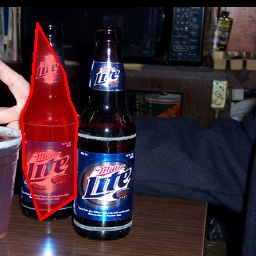}}\,\!\!
    \subfloat {\includegraphics[width=1.68cm,height=1.4cm]{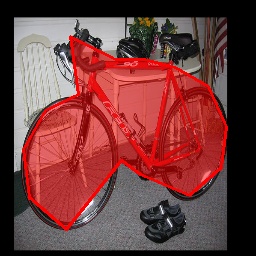}}\,\!\!
    \subfloat {\includegraphics[width=1.68cm,height=1.4cm]{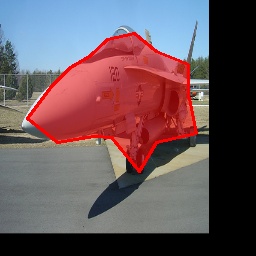}}\,\!\!
    \subfloat {\includegraphics[width=1.68cm,height=1.4cm]{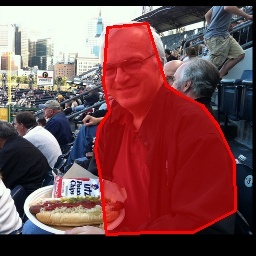}}\,\!\!
    \subfloat {\includegraphics[width=1.68cm,height=1.4cm]{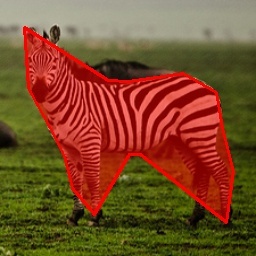}}\,\!\!
    \subfloat {\includegraphics[width=1.68cm,height=1.4cm]{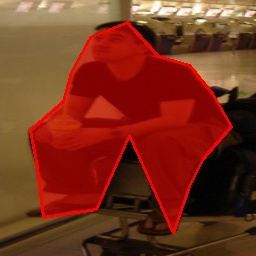}}\\[-4.7ex]

    \setcounter{subfigure}{-1}
    \subfloat {\raisebox{1.6em}{\rotatebox[origin=t]{90}{\scriptsize ESE-Seg \cite{xu2019}}}}\hspace{-0.01cm}\,\!
    \subfloat {\includegraphics[width=1.68cm,height=1.4cm]{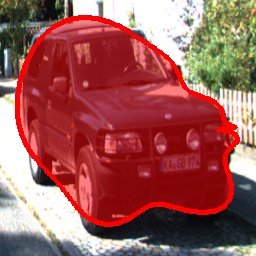}}\,\!\!
    \subfloat {\includegraphics[width=1.68cm,height=1.4cm]{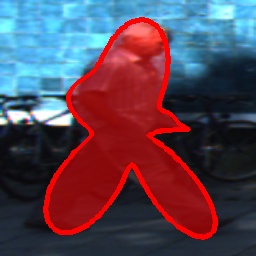}}\,\!\!
    \subfloat {\includegraphics[width=1.68cm,height=1.4cm]{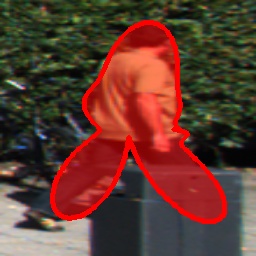}}\,\!\!
    \subfloat {\includegraphics[width=1.68cm,height=1.4cm]{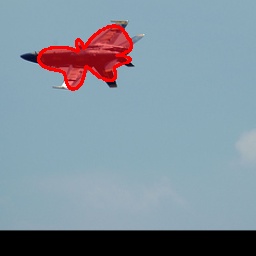}}\,\!\!
    \subfloat {\includegraphics[width=1.68cm,height=1.4cm]{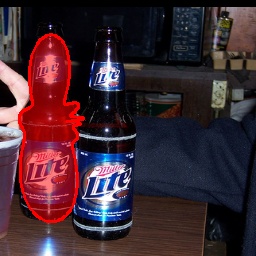}}\,\!\!
    \subfloat {\includegraphics[width=1.68cm,height=1.4cm]{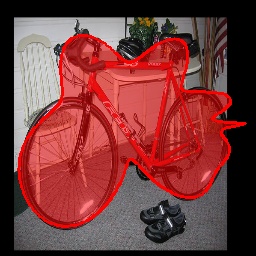}}\,\!\!
    \subfloat {\includegraphics[width=1.68cm,height=1.4cm]{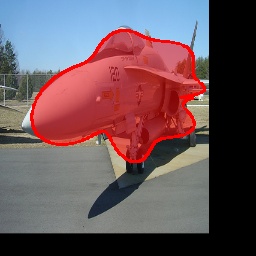}}\,\!\!
    \subfloat {\includegraphics[width=1.68cm,height=1.4cm]{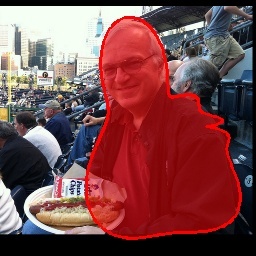}}\,\!\!
    \subfloat {\includegraphics[width=1.68cm,height=1.4cm]{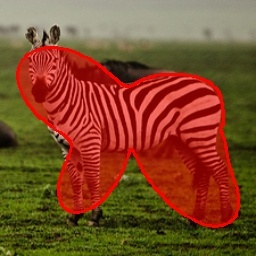}}\,\!\!
    \subfloat {\includegraphics[width=1.68cm,height=1.4cm]{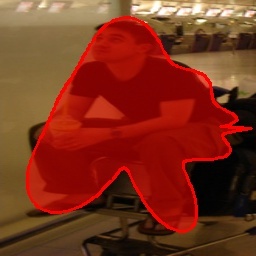}}\\[-4.7ex]

    \subfloat {\raisebox{1.6em}{\rotatebox[origin=t]{90}{\scriptsize Proposed}}}\hspace{-0.01cm}\,\!
    \subfloat {\includegraphics[width=1.68cm,height=1.4cm]{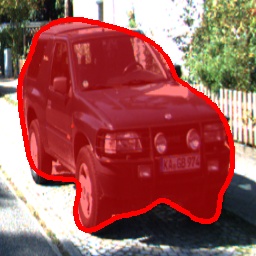}}\,\!\!
    \subfloat {\includegraphics[width=1.68cm,height=1.4cm]{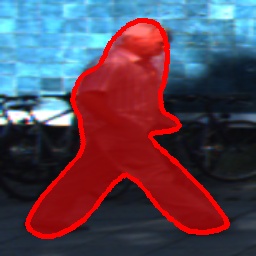}}\,\!\!
    \subfloat {\includegraphics[width=1.68cm,height=1.4cm]{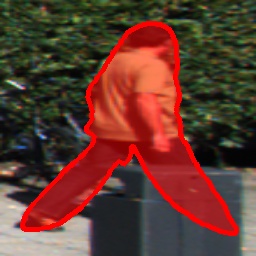}}\,\!\!
    \subfloat {\includegraphics[width=1.68cm,height=1.4cm]{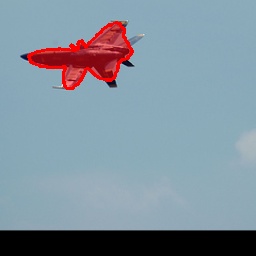}}\,\!\!
    \subfloat {\includegraphics[width=1.68cm,height=1.4cm]{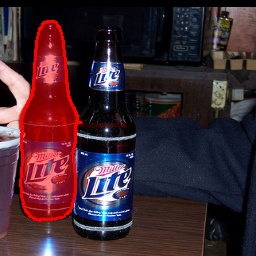}}\,\!\!
    \subfloat {\includegraphics[width=1.68cm,height=1.4cm]{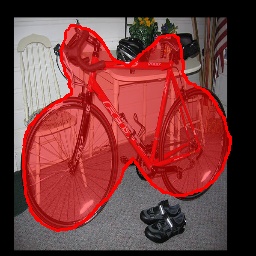}}\,\!\!
    \subfloat {\includegraphics[width=1.68cm,height=1.4cm]{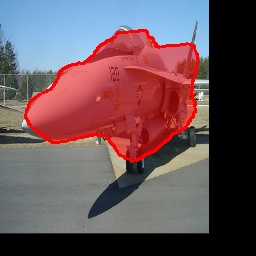}}\,\!\!
    \subfloat {\includegraphics[width=1.68cm,height=1.4cm]{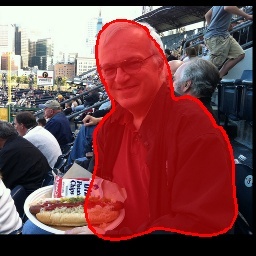}}\,\!\!
    \subfloat {\includegraphics[width=1.68cm,height=1.4cm]{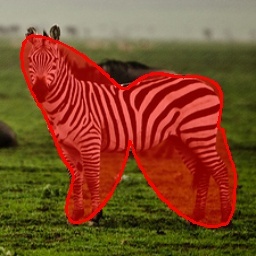}}\,\!\!
    \subfloat {\includegraphics[width=1.68cm,height=1.4cm]{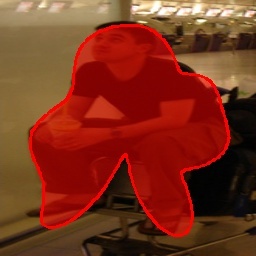}}\\[-4.7ex]

    \setcounter{subfigure}{-1}

    \subfloat {\raisebox{1.6em}{\rotatebox[origin=t]{90}{\scriptsize Ground-truth}}}\hspace{-0.02cm}\,
    \subfloat {\includegraphics[width=1.68cm,height=1.4cm]{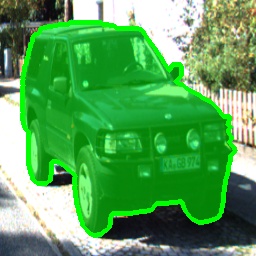}}\,\!\!
    \subfloat {\includegraphics[width=1.68cm,height=1.4cm]{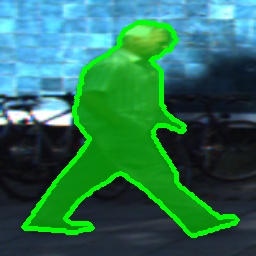}}\,\!\!
    \subfloat {\includegraphics[width=1.68cm,height=1.4cm]{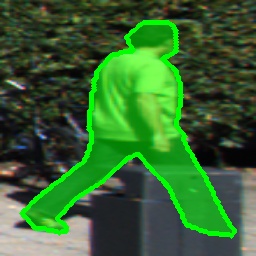}}\,\!\!
    \subfloat {\includegraphics[width=1.68cm,height=1.4cm]{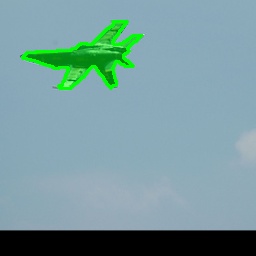}}\,\!\!
    \subfloat {\includegraphics[width=1.68cm,height=1.4cm]{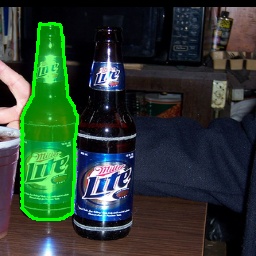}}\,\!\!
    \subfloat {\includegraphics[width=1.68cm,height=1.4cm]{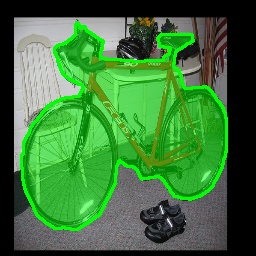}}\,\!\!
    \subfloat {\includegraphics[width=1.68cm,height=1.4cm]{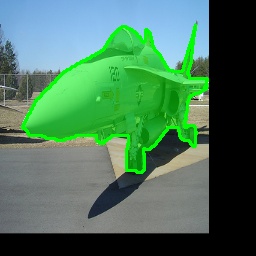}}\,\!\!
    \subfloat {\includegraphics[width=1.68cm,height=1.4cm]{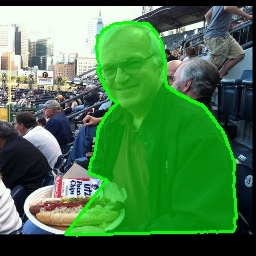}}\,\!\!
    \subfloat {\includegraphics[width=1.68cm,height=1.4cm]{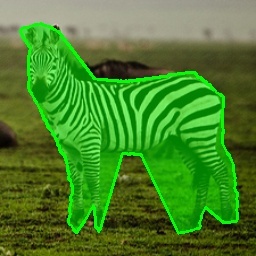}}\,\!\!
    \subfloat
    {\includegraphics[width=1.68cm,height=1.4cm]{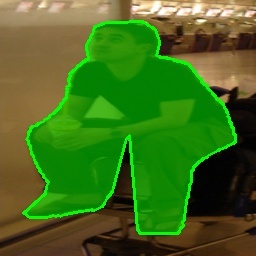}}\\

    \vspace*{-0.4cm}
    \caption
    {
        Qualitative comparison of boundary representations at $M=16$. The left three images are from KINS, the middle three from SBD, and the remaining four from COCO2017.
    }
    \label{fig:compare_approx_fig}
\end{figure*}

\subsection{Comparative Assessment}
\label{ssec:assess}

\noindent\textbf{Contour descriptors:}
It is desirable for contour descriptors to represent an object boundary compactly, as well as to reconstruct it faithfully. We compare the proposed eigencontours with the conventional contour descriptors \cite{xu2019, xie2020}. For contour description, centroidal profiles are used in PolarMask \cite{xie2020}, while polynomial fitting is performed to approximate the shape signature of a boundary in ESE-Seg \cite{xu2019}. In this test, the proposed eigencontours are determined for all instances in all categories in a training dataset.

For the quantitative assessment of contour descriptors, we employ the F-measure ($\cal{F}$) \cite{perazzi2016}. Specifically, bipartite matching is performed between the boundary points of a ground-truth contour and its approximated version. Then, the $\cal F$ score is defined as the harmonic mean of the precision ($\cal{P}$) and the recall ($\cal{R}$) of the matching results.

\begin{figure*}[t]
\vspace{-0.1cm}
  \centering
  \includegraphics[width=1\linewidth]{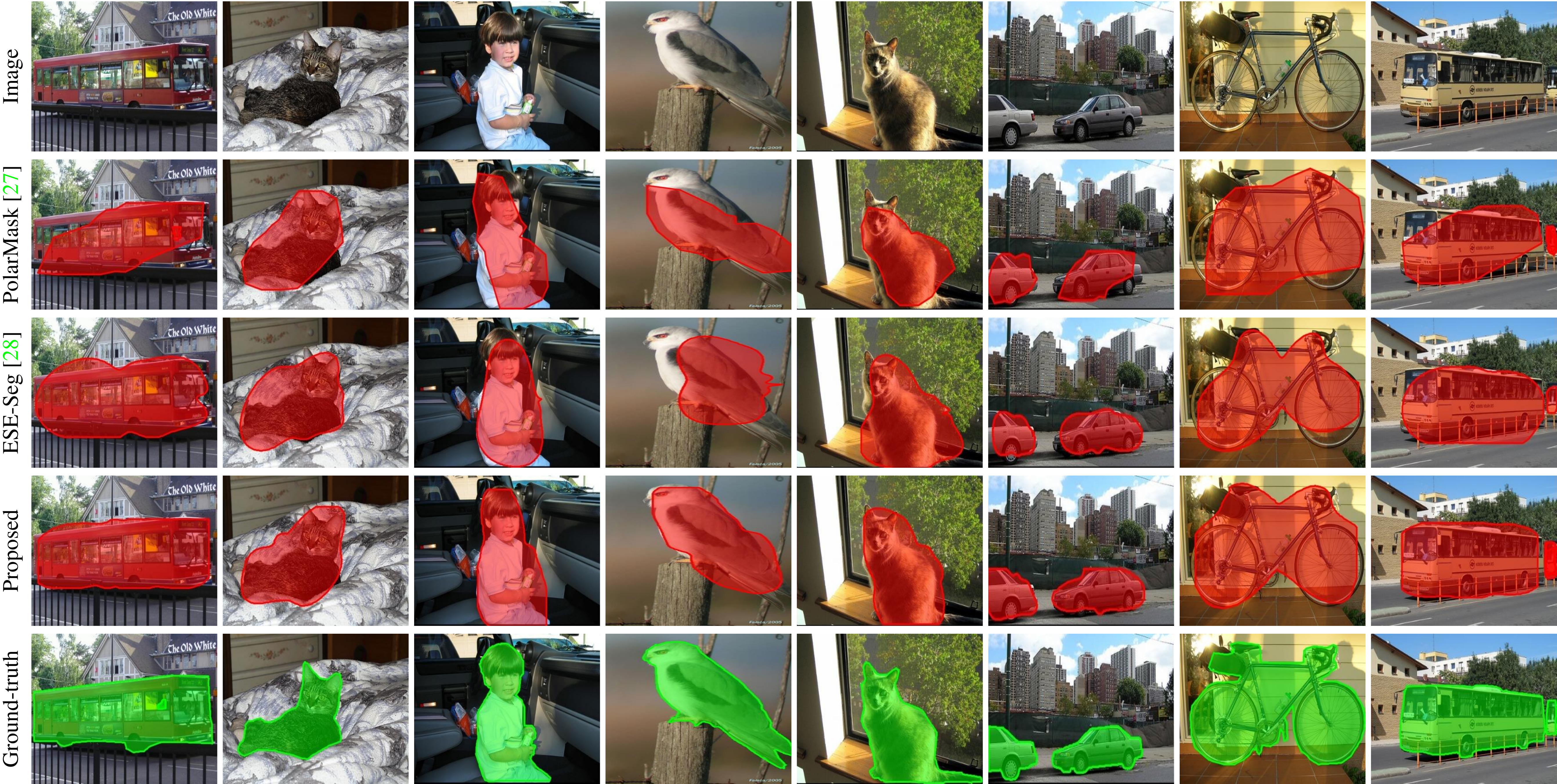}
  \caption{Comparison of instance segmentation results on the SBD dataset}
  \vspace{-0.2cm}
  \label{fig:Result_inst_seg}
\end{figure*}

Figure~\ref{fig:graph_fig} compares the $\cal{F}$ curves of the proposed eigencontours with those of  the conventional descriptors according to the dimension $M$ of the descriptors. In PolarMask, $M$ radial coordinates in a centroidal profile are sampled to describe a contour. In ESE-Seg, $M$ is the number of Chebyshev polynomial coefficients for approximating a contour. For all three datasets of KINS, SBD, and COCO2017, the proposed algorithm outperforms both PolarMask and ESE-Seg at every $M$. For KINS, the proposed algorithm achieves an $\cal F$ score higher than 0.9 at $M=24$, while the conventional ones need approximately double the dimension to yield a similar $\cal F$ score. For SBD, similar tendencies are observed. For COCO2017, containing diverse instances with complicated shapes, the instances require higher-dimensional description than those in KINS and SBD. However, the proposed algorithm is still superior to the conventional ones.

\begin{table}[t]\centering
    \renewcommand{\arraystretch}{1.0}
    \caption
    {
        AUC-$\cal{F}$ performances on KINS, SBD, and COCO2017.
    }
    \vspace*{-0.15cm}
    \resizebox{0.95\linewidth}{!}{
    \begin{tabular}[t]{+L{3.0cm}^C{1.6cm}^C{1.6cm}^C{1.6cm}}
    \toprule
    & KINS & SBD & COCO2017 \\
    \midrule
         PolarMask \cite{xie2020}      & 75.47 & 76.23 & 74.05\\
         ESE-Seg \cite{xu2019}             & 77.37 & 76.86 & 70.21\\
    \midrule
         Proposed                  & \bf{89.17} & \bf{86.51} & \bf{76.92}\\
    \bottomrule
    \end{tabular}}
    \vspace{-0.2cm}
    \label{table:result_fscore}
\end{table}

Table \ref{table:result_fscore} compares the area under curve performances of the $\cal{F}$ curves (AUC-$\cal{F}$) in Figure \ref{fig:graph_fig}. The proposed algorithm outperforms the conventional algorithms by significant margins on all datasets. In other words, the proposed algorithm represents object boundaries more faithfully than the conventional algorithms, when the same number of coefficients are used for the contour description.

Figure \ref{fig:compare_approx_fig} compares object boundaries approximated by the contour descriptors at $M=16$. PolarMask fails to reconstruct curved parts. ESE-Seg provides better results, but it blurs complicated parts, especially the leg boundaries in the second and third columns. In contrast, the proposed eigencontour descriptors represent the object boundaries more accurately and more reliably.
\begin{table}[t]\centering
    \renewcommand{\arraystretch}{1.0}
    \caption
    {
        Comparison of the clustering performances on the COCO2017 dataset at $M=16$ and $K=500$.
    }
    \resizebox{0.95\linewidth}{!}{
    \begin{tabular}[t]{+L{3.0cm}^C{1.6cm}^C{1.6cm}^C{1.6cm}}
    \toprule
    & $\cal{P}$ & $\cal{R}$ & $\cal{F}$ \\
    \midrule
         PolarMask \cite{xie2020}      & 28.59 & 22.67 & 25.13 \\
         ESE-Seg \cite{xu2019}             & 30.31 & 24.31 & 26.82 \\
    \midrule
         Proposed                  & \bf{30.88} & \bf{24.90} & \bf{27.40} \\
    \bottomrule
    \end{tabular}}
    \vspace{-0.1cm}
    \label{table:result_fscore_clustering}
\end{table}

\vspace*{0.1cm}
\noindent\textbf{Clustering in low-dimensional space:}
As mentioned in Section~\ref{ssec:formulation}, it is possible to cluster object contours in a lower-dimensional descriptor space and obtain contour centroids there. To validate the effectiveness of the clustering in the proposed eigencontour space, we compare the clustering performances of the proposed algorithm on the COCO2017 dataset with those of PolarMask and ESE-Seg. To this end, we employ each algorithm to approximate all training boundaries into $M$-dimensional descriptors and obtain $K$ centroids via $K$-means. Then, each contour in the dataset is matched with the nearest centroid, and the matching performance is computed in terms of $\cal{P}$, $\cal{R}$, and $\cal{F}$.

Table \ref{table:result_fscore_clustering} compares the performances at $M=16$ and $K=500$. The proposed algorithm yields the best results in all three metrics, which indicates that the proposed algorithm can process object contours more reliably in a low-dimensional space. Qualitative comparison results of the clustering are available in the supplemental document.

\begin{table}[t]\centering
    \renewcommand{\arraystretch}{1.0}
    \caption
    {
        Comparison of the $\text{AP}_{50}$, $\text{AP}_{75}$, and $\text{AP}_{\cal{F}}$ performances on the SBD validation dataset.
    }
    \resizebox{0.95\linewidth}{!}{
    \begin{tabular}[t]{+L{3.0cm}^C{1.6cm}^C{1.6cm}^C{1.6cm}}
    \toprule
    & $\text{AP}_{50}$ & $\text{AP}_{75}$ & $\text{AP}_{\cal{F}}$ \\
    \midrule
         PolarMask \cite{xie2020}      & 50.11 & 14.50 & 25.78 \\
         ESE-Seg \cite{xu2019}             & 52.14 & 20.48 & 27.37 \\
    \midrule
         Proposed                  & \bf{56.47} & \bf{29.35} & \bf{35.30} \\
    \bottomrule
    \end{tabular}}
    \vspace{-0.1cm}
    \label{table:result_instance}
\end{table}

\vspace*{0.1cm}
\noindent\textbf{Instance segmentation:}
Both PolarMask and ESE-Seg were proposed for instance segmentation. To localize each instance, these methods reformulate the pixelwise classification as the regression of an object contour. The proposed eigencontours are more effective for this instance segmentation task as well. To demonstrate this, as done in ESE-Seg, we adopt YOLOv3 \cite{redmon2018} as an object detector and modify its components. Given an input image, we predict an output map, in which each element contains an $M$-dimensional coefficient vector as well as the original YOLOv3 vector for bounding box regression and object classification. Then, we use the coefficient vector to linearly combine eigencontours to reconstruct the contour and shape mask of an object.  The supplemental document describes the implementation details and the training procedure.

\begin{figure}[t]
  \centering
  \setlength{\abovecaptionskip}{-0.01cm}
  \setlength{\belowcaptionskip}{-0.18cm}
  \includegraphics[width=1\linewidth]{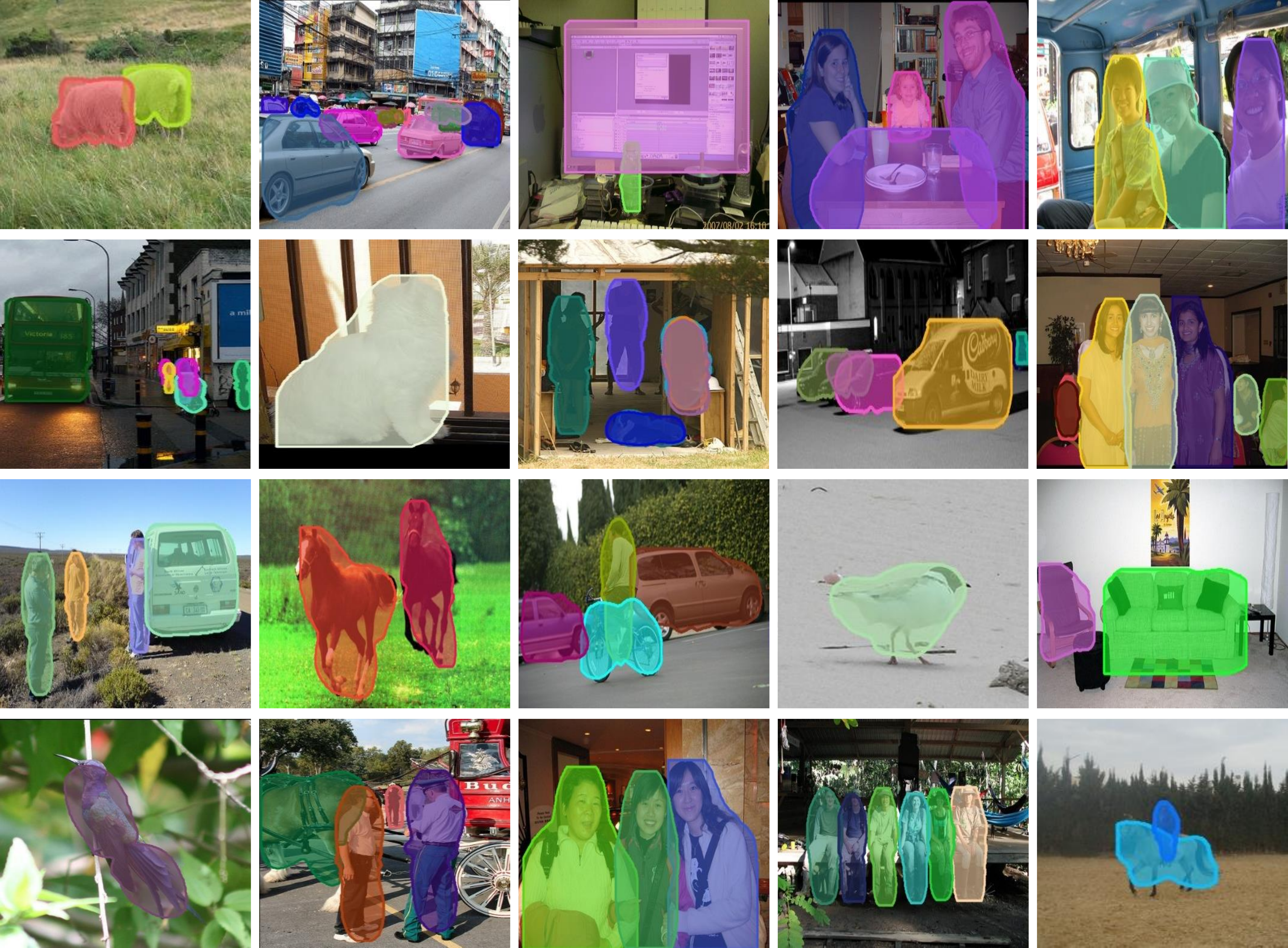}
  \vspace*{0.1cm}
  \caption{Instance segmentation results of the proposed algorithm on the SBD dataset.}
  \label{fig:result}
\end{figure}

Table \ref{table:result_instance} compares the instance segmentation results on the SBD validation dataset at $M=20$. The average precision (AP) performances, based on two intersection-over-union (IoU) thresholds of 0.5 and 0.75 and an $\cal{F}$ score threshold of 0.3, are reported. The proposed algorithm performs better than PolarMask and ESE-Seg in terms of all three metrics. Figure~\ref{fig:Result_inst_seg} shows boundary regression results. PolarMask and ESE-Seg fail to reconstruct object boundaries reliably. In contrast, the proposed algorithm  represents the boundaries more faithfully. Figure \ref{fig:result} shows more instance segmentation results.

\subsection{Analysis}
\label{ssec:ablation}

\vspace*{0.1cm}
\noindent\textbf{Dimension of eigencontour space ($M$):}
Table~\ref{table:ablation_m} lists the AUC-$\cal{F}$ performances of the proposed algorithm on the SBD validation dataset according to the dimension, $M$, of the eigencontour space. At $M=10$, the proposed algorithm yields poor scores, since object boundaries are too simplified and not sufficiently accurate. At $M=20$, it provides the best results. At $M=30$, it yields similarly good results. However, at $M=40$, the performances are degraded further, which indicates that a high-dimensional space does not always lead to better results. It is more challenging to regress more variables reliably. There is a tradeoff between accuracy and reliability. In this test, $M=20$ achieves a good tradeoff.

\vspace*{0.1cm}
\noindent\textbf{Categorical eigencontour space:}
The proposed eigencontours are data-driven descriptors, which depend on the distribution of object contours in a dataset. Thus, different eigencontours are obtained for different data. Let us consider two options for constructing eigencontour spaces: categorial construction and universal construction.
In the categorial construction, eigencontours are determined for each category in a dataset. In the universal construction, they are determined for all instances in all categories.

For the two options, $\cal{F}$ score curves are presented according to the dimension $M$ in the supplemental document. Table \ref{table:ablation_class} compares the area under curve performances of the $\cal{F}$ curves up to $M=18$. The categorial construction provides better performances than the universal construction, because it considers similar shapes in the same category only. In COCO2017, the gap between the two options is the smallest. This is because some object shapes are not properly represented due to occlusions and thus COCO2017 objects exhibit low intra-category correlation. In contrast, in KINS, whole contours are well represented because occluded regions are also annotated. Hence, the gap between the two options is the largest.

\vspace*{0.1cm}
\noindent\textbf{Limitations:}
The proposed eigencontours represent typical contour patterns in a dataset. Thus, if object contour patterns differ among datasets, the eigencontours for a dataset may be effective for that particular dataset only. To assess the dependency of eigencontours on a dataset, we conduct cross-validation tests between datasets in the supplemental document.

\begin{table}[t]\centering
    \renewcommand{\arraystretch}{1.0}
    \caption
    {
        The instance segmentation performances of the proposed algorithm on the SBD validation dataset according to the dimension of the eigencontour space ($M$).
    }

    \vspace*{0.1cm}
    \resizebox{0.95\linewidth}{!}{
    \begin{tabular}[t]{+C{2.9cm}^C{1.8cm}^C{1.8cm}^C{1.8cm}}
    \toprule
    $M$ & $\text{AP}_{50}$ & $\text{AP}_{75}$ & $\text{AP}_{\cal{F}}$ \\
    \midrule
        10     & 49.96 & 26.19 & 30.27 \\
        20     & \bf{56.47} & \bf{29.35} & 35.30 \\
        30     & 55.85 & 28,89 & \bf{36.15} \\
        40     & 54.45 & 22.42 & 32.21 \\
    \bottomrule
    \end{tabular}}
    \vspace{0.1cm}
    \label{table:ablation_m}
\end{table}

\begin{table}[t]\centering
    \renewcommand{\arraystretch}{0.95}
    \caption
    {
        Comparison of the AUC-$\cal{F}$ performances of categorical and universal eigencontours, $M \in [3, 18]$.
    }
    \vspace*{0.1cm}
    \resizebox{0.92\linewidth}{!}{
    \begin{tabular}[t]{+L{3.0cm}^C{1.6cm}^C{1.6cm}^C{1.6cm}}
    \toprule
    & KINS & SBD & COCO2017 \\
    \midrule
         Universal       & 64.65 & 60.22 & 47.19 \\
         Categorical  & 67.67 & 62.37 & 48.77 \\
    \bottomrule
    \end{tabular}}
    \label{table:ablation_class}
\end{table}


\section{Conclusions}
We proposed novel contour descriptors, called eigencontours, based on low-rank approximation. First, we constructed a contour matrix containing all contours in a training set. Second, we approximated the contour matrix, by performing the best rank-$M$ approximation. Third, we represent an object boundary by a linear combination of the $M$ eigencontours. Experimental results demonstrated that the proposed eigencontours can represent object boundaries more effectively and more faithfully than the existing methods. Moreover, the proposed algorithm yields meaningful instance segmentation performances.

\section*{Acknowledgements}
This work was supported by the National Research Foundation of Korea (NRF) grants funded by the Korea government (MSIT) (No.~NRF-2021R1A4A1031864 and No.~NRF-2022R1A2B5B03002310).

\clearpage

{\small
\bibliographystyle{ieee_fullname}
\bibliography{2022_CVPR_WHPARK}
}

\end{document}